\documentclass[default,iicol]{sn-jnl}

\jyear{2021}%

\theoremstyle{thmstyleone}%
\theoremstyle{thmstyletwo}%

\theoremstyle{thmstylethree}%

\begin{document}

\title{Leveraging systems' non-linearity to tackle the scarcity of data in the design of Intelligent Fault Diagnosis Systems}


\author*[1,2]{\fnm{Giancarlo} \sur{Santamato}}\email{giancarlo.santamato@santannapisa.it, andreamattia.garavagno@santannapisa.it}
\equalcont{These authors contributed equally to this work.}

\author[1,2]{\fnm{Andrea Mattia} \sur{Garavagno}}
\equalcont{These authors contributed equally to this work.}

\author[1,2]{\fnm{Massimiliano} \sur{Solazzi}}\email{massimiliano.solazzi@santannapisa.it}

\author[1,2]{\fnm{Antonio} \sur{Frisoli}}\email{antonio.frisoli@santannapisa.it}

\affil*[1]{\orgdiv{Institute of Mechanical Intelligence}, \orgname{Scuola Superiore Sant'Anna}, \orgaddress{\street{via Alamanni 13b}, \city{Ghezzano}, \postcode{56010}, \state{Pisa}, \country{Italy}}}

\affil[2]{\orgdiv{Department of Excellence in Robotics \& AI}, \orgname{Scuola Superiore Sant'Anna}, \city{Pisa}, \country{Italy}}


\abstract{Deep Transfer Learning (DTL) allows for the efficient building of Intelligent Fault Diagnosis Systems (IFDS). On the other hand, DTL methods still heavily rely on large amounts of labelled data. Obtaining such an amount of data can be challenging when dealing with machines or structures faults. This document proposes a novel approach to the design of vibration-based IFDS using DTL in condition of strong data scarcity. A periodic multi-excitation level procedure leveraging intrinsic non-linearities of real-world systems is used to produce images that can be conveniently analysed by pre-trained Convolutional Neural Networks (CNNs) to diagnose faults. A new data visualization method and its augmentation technique are proposed in this paper to tackle the typical lack of data encountered during the design of IFDS. Experimental validation on a railway pantograph structure provides effective support for the proposed method.}


\maketitle

\footnotetext{\textbf{Notice:} This is a preprint version of a paper published in \textit{Nonlinear Dynamics}. Please cite the final version: Santamato, G., Garavagno, A.M., Solazzi, M., Frisoli, A. Leveraging systems’ non-linearity to tackle the scarcity of data in the design of intelligent fault diagnosis systems. \textit{Nonlinear Dynamics} 112, 16153–16166 (2024). \url{https://doi.org/10.1007/s11071-024-09864-6}}

\section{Introduction}\label{sec:introduction} 
Deep Transfer Learning (DTL) is enabling the rise of the next-generation machine and structural health monitoring systems \cite{avci2021review}. The advantage of Deep Learning (DL) in feature representation combined with the ability to reuse the knowledge extracted from large datasets, typical of Transfer Learning, allows for the efficient building of Intelligent Fault Diagnosis Systems (IFDS). 

On the other hand, DTL methods heavily rely on large amounts of labelled data.
Obtaining such an amount of data can be challenging when dealing with real-world machines or structures. On the other side, most faults are rare and irreproducible in laboratory facilities. Therefore researchers focus on systems of industrial interest which can be tampered with easiness and low cost as small frames or also rotating machinery like gearboxes and rolling bearings. Compared to the multiplicity of IFDS presented in the scientific literature, few works propose techniques to tackle the typical data scarcity encountered when designing IFDS in real-world scenarios. Hence, tackling the data lack is still an open issue \cite{li2022perspective}. 

To overcome the aforementioned deficiency, we exploited the typical structures' nonlinearity to design a novel data visualization method and its augmentation technique. The proposed approach belongs to the so-called non-parametric techniques, meaning that it only relies on features that are extracted directly from raw data with no need of physical models. In particular, this method is suitable for structural dynamic tests where the excitation level can be controlled externally, as for example in modal experimental testing where the amplitude of the applied excitation can be tuned through external actuators. 

The underlying idea is that in the presence of nonlinearity, the frequency response characteristics of the structure are dependent on the excitation level.   
What is more, also the damage signature in the frequency domain is modulated by the amplitude of the external force.
Consequently, for a given fault type, detection may be enhanced at low or high excitation levels, as demonstrated in the literature \cite{santamato2024investigating}.
Therefore, the increased information derived from the excitation level is exploited to produce 2D images.

In this work, the system response is conveyed through the Frequency Response Function (FRF), represented as the dB-magnitude of the receptance function (displacement-over-excitation).
The FRFs acquired at different excitation levels are collected in a single colour map, thus producing a spectrogram-like image where the axis of time is substituted by the axis of the excitation level. 
Such a representation enables a novel custom procedure of data augmentation, where different acquisitions of the same fault are mixed to create a multitude of new realistic colour maps, thus leveraging the excitation levels multiplicity.

Besides, the proposed visualization method enables the well-known power of pre-trained Convolutional Neural Networs (CNNs) on ImageNet \cite{russakovsky2015imagenet}. 
In contrast to the widely used General Adversial Networks (GANs) for data augmentation \cite{shao2019generative, li2022multi, gao2020data, fu2020novel, lu2021lightweight, guo2022data}, our custom technique does not require any additional training. No existing dataset is required to train from scratch an Artificial Intelligence (AI) model able to generate synthetic data. Our technique can be applied in cases of high data scarcity, as in the case of more traditional data augmentation techniques like Data Resampling (DSR) \cite{hu2019data, hu2020simple}, slicing \cite{zhang2018deep} and noise injection.
The obtained images can be further augmented by applying general data augmentation techniques for images like Mixup \cite{zhang2017mixup}, Cutmix \cite{yun2019cutmix}, RandAugment \cite{cubuk2020randaugment} and Random Erasing \cite{zhong2020random}.
Lastly, the proposed approach is also convenient since it relies only on simple Fast-Fourier-Transform routines with no need of complex models and computational burden. 

The feasibility of the method is investigated experimentally through dedicated structural dynamics tests performed on a railway pantograph mechanism, as an example of inherently nonlinear system owing to dry-frictional nonlinearity \cite{santamato2023detecting}. Two damage scenarios, namely the loss of a member connectivity at a bolted connection and a tampering of the artificial damping, have been tested while the excitation varied on seven levels, distributed over one order of magnitude and stimulating stuck and stick-slip regimes.
Therefore we show how our IFDS is able to correctly classify the undamaged condition and the fault scenarios with a relatively low-size data set.

The paper is structured as follows: a survey of the literature delving with non-parametric fault diagnosis is reviewed from classical approaches to AI methods in Sec. \ref{sec:related_work}. The proposed approach is presented in Sec. \ref{sec:proposed_method} as long with the assumed hypotheses and assumptions. Sec. \ref{sec:experiments} is devoted to describing the experimental tests performed on a railway pantograph and to discuss the fault diagnosis performance achieved through our IFDS. Sec. \ref{sec:conclusions} synthesizes the work and highlights the future developments.

\section{Related Work}\label{sec:related_work}
Vibration-based IFDSs are often found in the scientific literature \cite{avci2021review}. In nonparametric methods, the damage features are extracted directly from raw data without the need of physical models. Typically, the stimulus response is treated as a numerical series in the time or frequency domain and processed either by a classifier or by an outlier detector to evaluate the status of the structure. Both big data and classical techniques are used to diagnose faults. 

Classical approaches often utilized the experimental estimate of the Frequency Response Function (FRF) as a numerical series \cite{kopsaftopoulos2010vibration}. In the Waveform Chain Code (WCC) approaches, the damage index is obtained by summing up the contributions of the absolute difference between the first and/or second derivatives of the FRFs series in the damaged and undamaged conditions \cite{biswas1994modified}. The FRF-based curvature method was proved to successfully identify and locate multiple damage sites in a steel beam \cite{porcu2019effectiveness}.
The WCC can be also performed on Principal Component Analysis (PCA)-reduced FRFs. In \cite{chen2019waveform,chen2020unsupervised} k-means unsupervised clustering with unlabeled data allowed to separate with high-accuracy damage conditions such as screws removed or loosened. 
The Interpolation Damage Detection Method (IDDM) was utilized on bridges and frame structures to detect early damages and location, in the presence of noise and environmental changes \cite{limongelli2010frequency,limongelli2011interpolation}. The FRF Similarity Metric (FRFSM) was proposed in \cite{lee2018metric} as a statistical method to compare the dB-scale magnitude of two FRFs, based on the probability density function of normal distribution in the frequency domain. 

Big data techniques like Support Vector Machines \cite{ali2015application, zheng2023composite}, decision trees \cite{li2016fault}, Random Forests (RF) \cite{wei2021intelligent}, k-nearest neighbors (KNN) \cite{uddin2016distance} and dictionary-based learning \cite{zhao2019weighted, wang2019supervised}, as well as deep learning ones like Auto Encoders (AEs) \cite{shao2018novel}, Deep Belief Networks (DBN) \cite{shao2018rolling}, Recurrent Neural Networks (RNNs) \cite{zhang2021fault, fang2021ans}, Spiking Neural Networks (SNNs) \cite{zuo2022multi}, General Adversial Networks (GANs) \cite{zhou2023deep}, one-dimensional CNNs \cite{li2020gear} and bi-dimensional CNNs \cite{an2022rolling} are also used to diagnose faults, often in combination with classical techniques. 

In the case of IFDS based on bi-dimensional CNNs, acquired time series are transformed into images. Techniques like short-time Fourier transform (STFT) \cite{zhang2023intelligent, verstraete2017deep, pandhare2019convolutional, wen2019new} and Wavelet Transform (WT) \cite{islam2019automated, gao2018fault, he2022bearing, guo2018novel} are widely applied to produce spectrogram images. Time series are also visualized as images in polar coordinates applying the principle of symmetrized dot pattern \cite{sun2022bearing}, or as bi-dimensional Kurtograms \cite{prosvirin2018bearing}. Sometimes the samples directly become the pixel of a grayscale image \cite{zhang2020new}, or time series coming from different sensors compose a bi-dimensional matrix \cite{xia2017fault}. Moreover, a visualization technique based on the PE has been proposed \cite{landauskas2020permutation}. It uses non-uniform embedding of the vibration signal into a delay coordinate space with variable time lags to produce images.

Nevertheless, many engineering structures are inherently nonlinear due to complex joints and interfaces, such as the presence of dry-friction at the connections. Consequently, the characteristics of the structure not only in the undamaged, but also in the damaged condition can be found dependent on the excitation amplitude, based on whether the external excitation breaks the joints friction. However, few existing algorithms consider nonlinear structural behaviour in the reference and damaged states \cite{hou2021review}.
In this regard, \cite{santamato2024investigating} demonstrated numerically and experimentally that the damage signature is dependent on the friction-over-excitation ratio also in a statistical sense. The relevance of the input excitation level was supported by numerous experimental tests, performed on a railway pantograph structure, and discussed for several kinds of defects
\cite{santamato2023detecting} through the analysis of the FRF through classical statistical analysis based on a p-value formulation.

\section{The proposed approach to Intelligent Fault Diagnosis} \label{sec:proposed_method}
The proposed approach uses a vibration-based multi-level excitation procedure to produce images representing the FRF of the structure under test, which is stimulated with an external excitation $q(t)$, artificially generated by a dedicated equipment, and which can be eventually combined with environmental sources.

\subsection{Hypotheses and assumptions}
The application of our methodology relies on a fundamental assumption, that is inherent nonlinearity, which implies that the structural response is dependent on the vibration amplitude both in the undamaged and damaged condition.
Despite the assumption of nonlinear behavior, the eligibility of the FRF still holds for the purpose of intelligent fault diagnosis.

When nonlinearity is present, the spectral content of the response $x(t)$ under a harmonic excitation at a frequency $\rho_{d}$ is in general spread over a multitude of frequency bins other than the excitation frequency $\rho_{d}$.
Nonetheless, in many cases, like Coulomb friction nonlinearity, the magnitude of these spurious harmonics is found negligible compared to the excitation frequency.
Consequently, the response can be expressed in the frequency domain as follows:
\begin{equation}
    X\left(\rho_d\right) = \mathcal{F}\left[x(t)\right]_{\rho = \rho_d}
\end{equation}
where $\mathcal{F}$ stands for the Fourier transform. In other words, only the principal harmonic $\rho_{d}$ can be retained from the entire transform signal.

Such a linearization procedure still holds when the input takes the form of a white-noise signal, i.e. random sequence or chirp waveform, which represent most common types of excitation in the experimental routines.
Hence, a frequency response function can be estimated in the classical non-parametric form, as the ratio of the Fourier transform of the input and output data record \cite{schoukens1998parametric}:
\begin{equation} \label{eqn:frf}
    FRF = \frac{\mathcal{F}\left[x\left(t\right) \right]}{\mathcal{F}\left[q(t)\right]}
\end{equation}

Yet, information about nonlinearity can be retained by carrying out several tests at different levels of the input excitation $Q_{0}$ (root-mean-squares, instantaneous amplitude etc.) \cite{worden2002nonlinearity,ewins2009modal,noel2017nonlinear}.
A further assumption is that the input excitation $Q_{0} $ can be tuned at least on two levels.

\subsection{The methodology}
\begin{figure*}[h!]
\centering
\includegraphics[width=\textwidth]{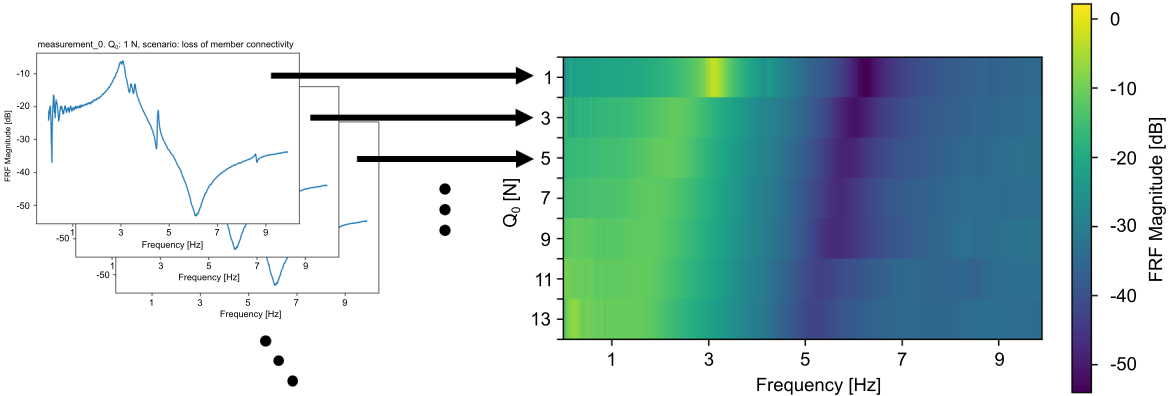}
\caption[caption]{A figure of the proposed visualization method. The Frequency Response Functions (FRF) acquired at each excitation level are on the rows of the colour map. The excitation levels are reported on the $y$-axis, while the frequencies on the $x$-axis. The corresponding magnitude is reported in the colour map legend. Yellowish colours indicate peaks of the FRF, while blueish colours indicate minima.}
\label{fig:data_representation_method}
\end{figure*}

Let us consider the structure under test to be excited at a given level $Q_{01}$. In experimental dynamics routines, structural excitation is repeated $K$ times for the sake of statistics, due to the presence of uncertainties. 

\begin{figure}
\centering
\includegraphics[width=0.45\textwidth]{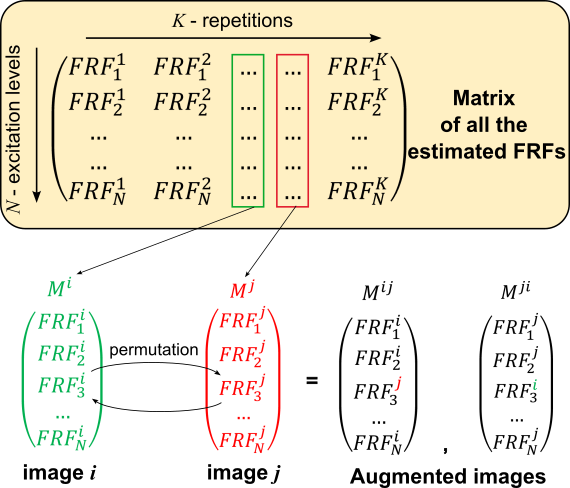}
\caption[caption]{A graphical representation of the concept behind the proposed data augmentation technique. Two matrices, $M^{i}$ and $M^{j}$ are extracted from the whole matrix of FRFs estimated in a given health condition. The permutation of two frequency response $FRF^{i}_{3}$ and $FRF^{i}_{3}$, at the same excitation level $Q_{03}$, creates two additional matrices $M^{ij}$ and $M^{ji}$ that will be plotted as a colour map.}
\label{fig:data_augmentation_technique}
\end{figure}

Consequently, for the given level $Q_{01}$, we have $K$ estimates of the frequency response function $FRF_{1}^{j}$ with $j = 1,...,K$, that can be stored in the first row of the matrix shown in Fig. \ref{fig:data_augmentation_technique}. 
Traditional noise mitigation techniques, like the p-Welch algorithm \cite{welch1967use}, exploit these repetitions to derive an averaged FRF, meaning that $K$ subsequent repetitions contribute to obtain one estimate of the frequency response.

The same procedure is repeated with several excitation levels $Q_{01}, ... Q_{0N}$, where $N$ is the number of chosen excitation levels, with $N \geq 2$. In the end, we will have a whole matrix $M$ collecting all the FRF-estimates, with $K$-columns (number of repetitions), and $N$-rows (number of excitation levels).

Each column of the matrix  $M$ contains the samples of the FRF and it can be plotted in the form of a colour map, as shown in Fig. \ref{fig:data_representation_method}, where the excitation levels are on the $y$-axis, while the frequencies on the $x$-axis. Lastly, the dB-magnitude of the frequency responses is shown in the colour map legend.

At this point, let us consider to extract two generic columns, $M^{i}$ and $M^{j}$ from the whole matrix $M$, and to permute one row, as shown in Fig. \ref{fig:data_augmentation_technique}. Such a swapping operation generates two additional columns $M^{ij}$, and $M^{ji}$ that can be converted in new images, and hence augmenting the available data.
Doing all the swaps allowed with $N$ excitation levels and $K$ repetitions we obtain $K^{N}$ images, thus augmenting the dataset of a factor ${K^{N}}/{K}$.

The proposed method has been tested and performed on each condition of the structure under test, including undamaged condition and each damage scenarios, as it will be discussed in Sec. \ref{sec:experiments}.
Besides, the introduction of the excitation level axis exploits the intrinsic non-linearities of real-world systems. Moreover, the power of pre-trained CNNs on Imagenet \cite{russakovsky2015imagenet} is used to analyze such non-linearities in order to diagnose faults, enabling the usage of DTL in the design of IFDS, even in strong data scarcity conditions.

\section{Experimental investigation}
With the purpose of validating the proposed methodology, we arranged an experimental setup devoted to fault diagnosis tests on a real structure.

\label{sec:experiments}
\subsection{Experimental setup and damage scenarios}
\begin{figure*}[htb]
\centering
\includegraphics[width=2
\columnwidth]{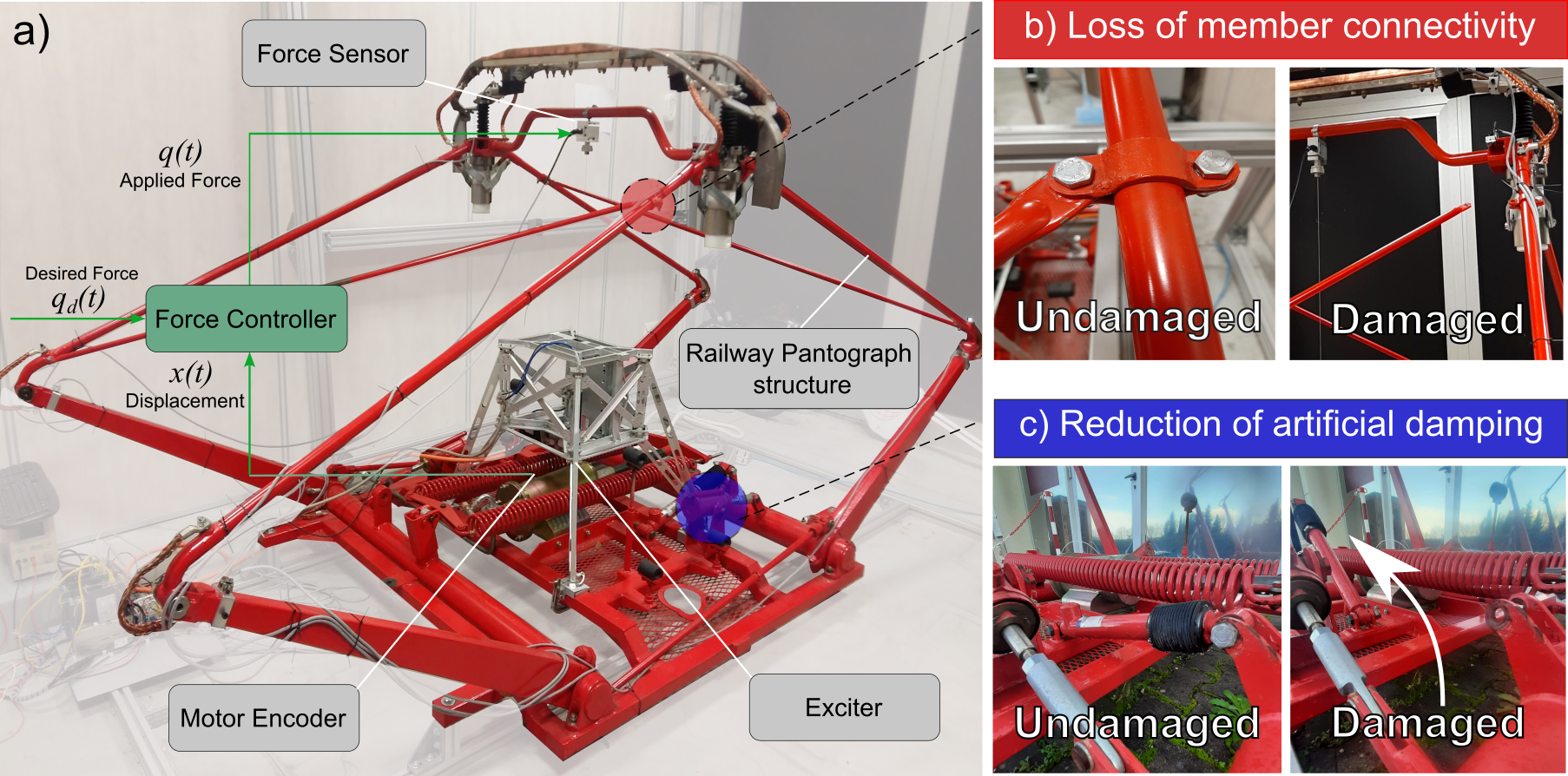}
	\caption{Experimental setup. (a) The structure under test - a railway pantograph - in the test configuration.  (b) Detail of the bolted connection that has been removed  to simulate the loss of member connectivity. (c) Detail of the hydraulic dashpot that has been removed to simulate the reduction of artificial damping} \label{fig:exp_setup}
\end{figure*}

The experimental setup, shown in Fig. \ref{fig:exp_setup}(a), consists of:
\begin{itemize}
    \item a real-scale railway pantograph, as an example of a structure with several dry-friction joints (base encumbrance: 1800x1350 $\text{mm}^2$; height of the pan head: 1500 mm, driven point mass: 35 kg);
    \item a custom exciter, endowed with a stinger actuator, able to set the level of the input excitation at a desired value within the range 1 - 13 N \cite{santamato2020lightweight};
    \item an analog force sensor at the driven-point, embedded with a bending parallelogram amplifier, providing a resolution of $5\cdot10^{-3}$ N;
    \item a digital encoder with a resolution of $5\cdot10^{-2}\,$ mm;
    \item a 16-bit DAC board and a communication module based on the EtherCAT protocol.
\end{itemize}

The railway pantograph is a linkage mechanism with parallel kinematics whose end-effector (the \textit{pan head}) is located at the top of the structure. Pantographs are specifically designed for electrical railways to supply energy from the contact between the pan head and the high-voltage line. Nevertheless, pantographs are prone to several faults, and their inspection is of crucial importance for the efficiency, safety and continuity of the rail service.
As shown in Fig. \ref{fig:exp_setup}(a), the exciter applies a dynamic force $q(t)$ along the vertical direction, measured by the force sensor, while the measured displacement $x$ of the actuation is admitted as the structural response. 

The desired excitation signal is generated by the controller commanding the actuation to generate a linear chirp waveform $q_{d}(t)$ whose instantaneous amplitude ${Q}_0$ is controlled by a closed-feedback loop. The objective of the control is to keep the instantaneous amplitude of the applied force close to ${Q}_0$ and constant at each frequency bin even in the presence of disturbances (actuator friction and inertia).

For this work, seven distinct values of the excitation amplitude ${Q}_0$ have been considered, ranging from 1 N to 13 N, i.e. $N$ = 7. The lowest level was chosen based on the force resolution of the actuation. Indeed, below 1 N, some excitation is still possible but the measurements are affected by a high signal-to-noise ratio. Besides, also the input-output coherence at the resonances drops at about 0.5, meaning poor correlation. On the other side, the highest excitation level is set at the power limit of the actuation. 

All signals are measured with 1000 Hz as the sampling rate. Based on the characteristics of the pantograph and the exciter, the frequency band of the excitation has been limited to the interval 0 - 10 Hz with a frequency resolution of 0.05 Hz.  
Six repetitions of the dynamic tests have been performed in each scenario ($K$ = 6).

The pantograph structure endorses several pinned joints with no-lubrication. Consequently, under external excitation, each joints can experience a stuck or a stick-slip regime, based on the local equilibrium forces. 
Previous studies already demonstrated the nonlinear behavior of the pantograph \cite{santamato2023detecting}, through the analysis of the FRF estimated for different values of the input excitation. In particular, the overall dry-friction level has been estimated at the driven-point through hysteresis tests, showing a value of hysteretic force equal to 15 N in the absence of other dissipation contributions. It comes that the friction-over-excitation ratio investigated in this study ranges from 1.15 to 15.

In the first explored scenario, we assume the pantograph to be undamaged, and stimulated with the routine described in Sec. \ref{sec:proposed_method}.
Afterward, tests have been repeated with the same routine for two simulated damaged scenarios. In particular, we removed the bolted connection of Fig. \ref{fig:exp_setup}(b) with the aim of reproducing the loss of member connectivity. 
It should be considered that bolt removal at this location has no crucial impact on the global integrity due to structural redundancy. Indeed, a twin connection is present on the opposite side of the pantograph. Hence the main effect of this fault is introducing in the explored spectrum the degrees-of-freedom of the diagonal, which can be considered as a local and subtle alteration to structural integrity. 
In the second damage scenario, we removed the artificial damper, shown in Fig. Fig. \ref{fig:exp_setup}(c), by unscrewing one of its pinned connections. In this faulty condition, the overall damping capability of the structure is reduced by 50\% (\cite{santamato2023detecting},\cite{santamato2019lightweight}) and the only damping source is due to joints friction.

\subsection{Analysis of the FRFs for multiple levels of the input excitation}
For the present study, the FRF was estimated for seven distinct levels of the input excitation: $\left[1, 3, 5, 7, 9, 11, 13\right]$ N.
The classical periodogram method \cite{welch1967use} was applied for the estimate in Eq. \ref{eqn:frf}.

\begin{figure*}
\centering
\includegraphics[width=\textwidth]{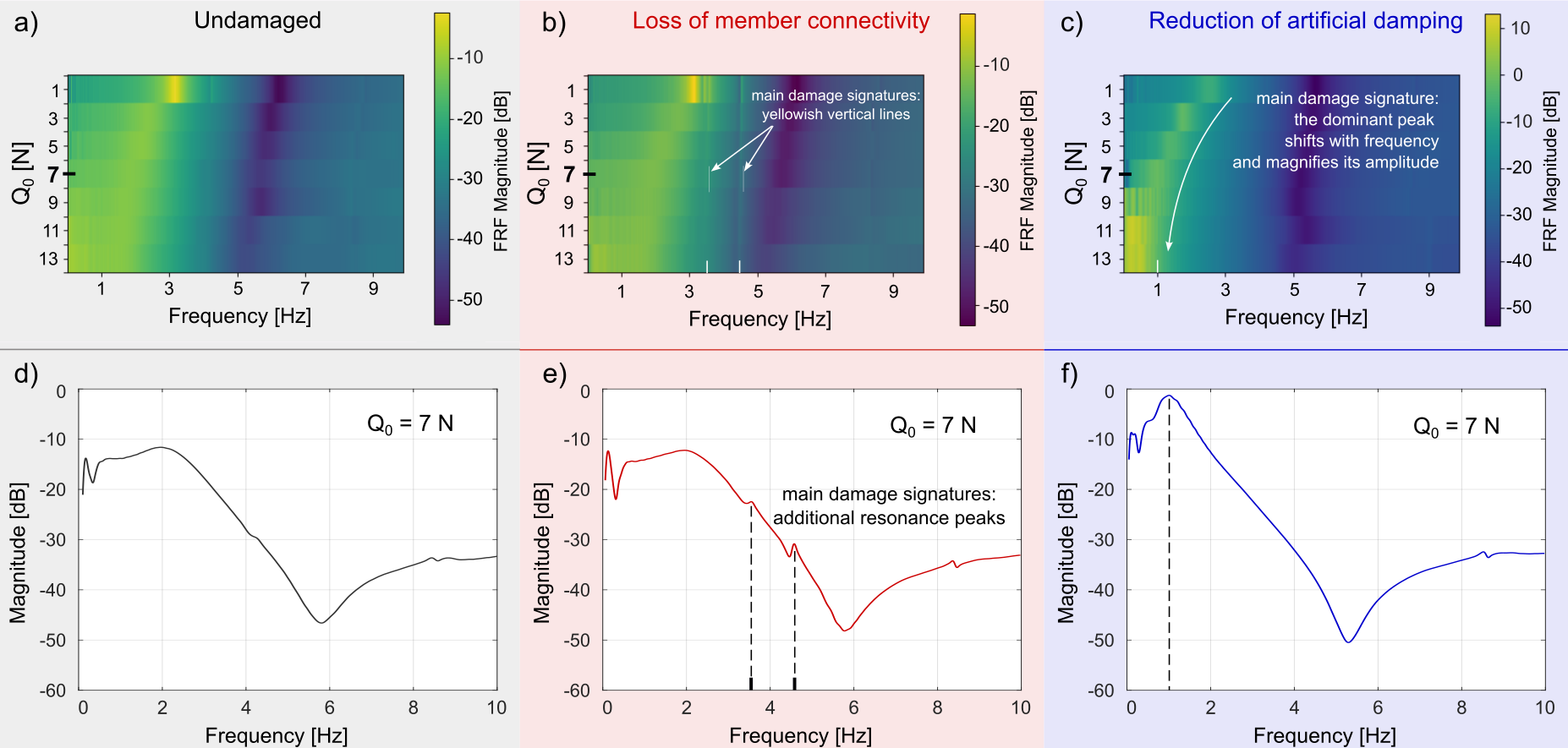}
\caption[caption]{Colour maps of the FRFs in the three explored scenarios: undamaged (a), loss of bolt connection (b), and reduction of damper (c). For each scenario, the FRF are reported below for a value of the external exctiation $Q_0$ equal to 7 N.}
\label{fig:comparison_FRF}
\end{figure*}

In the undamaged condition, Fig. \ref{fig:comparison_FRF}(a), the FRF is characterized by a dominant resonance peak and one zero.
At the frequency of the dominant peak, the response is magnified, as shown by the yellowish values of the contour map scale. On the contrary, at the frequency of the zero, the structure exhibits no displacement under the effect of the external excitation and the response on the spectrogram is blueish. By increasing the excitation level, both peaks are progressively shifted towards lower frequencies while their amplitude gets more damped.
This typical response function is better emphasised in Fig. \ref{fig:comparison_FRF}(d) showing the FRF for an excitation level of 7 N which is collocated at the mean of the explored interval. From such a detail, it also emerges the presence of a further residual peak, at around 4 Hz. 

The loss of member connectivity, Fig. \ref{fig:comparison_FRF}(b), introduces two additional resonance peaks which are reflected in the spectrogram by two vertical lines at 3.6 Hz, and 4.6 Hz, respectively. 
In the absence of the bolt, the diagonal bar behaves like a cantilever, that is free to oscillate with respect to the driven-point. First, a bending oscillation happens with a substantially vertical component of the bar tip, at 3.6 Hz, while a mainly lateral oscillation is shown in the second mode, at 4.6 Hz.
By increasing the excitation level, the sharpness of the damage-induced resonances is sensibly reduced. 
These damage-induce peaks are clearly visible in the detail of Fig. \ref{fig:comparison_FRF}(e).
As a consequence, the difference between the damaged and the undamaged FRF decreases with the excitation level by a factor of 10. 

The reduction of artificial damping, Fig. \ref{fig:comparison_FRF}(c), implies a shift of the dominant peak with the frequency and an increase of the amplitude. Besides, increasing the excitation level, the FRF exhibits a shifting peak (from 2.6 Hz to 0.5 Hz) and an increase of the amplitude of about 16 dB. Consequently, the difference between the damaged and the undamaged FRF is enhanced by a factor of 10, increasing the excitation level ratio.
The different profile of the FRF in this damage scenario is clearly visible in the detail of Fig. \ref{fig:comparison_FRF}(f), where the dominant peak achieves a value near 0 dB, at a frequency of 1 Hz.

\subsection{Design and test of the Intelligent Fault Diagnosis System}

To gather the data needed to design the IFDS, an acquisition campaign consisting of three acquisitions for each damage scenario at the seven excitation levels has been performed. The resulting FRFs have been plotted as in Fig. \ref{fig:data_representation_method}, obtaining three images per scenario. The resulting images have been augmented according to the technique proposed in Sec. \ref{sec:proposed_method}, retaining three of the six repetitions, i.e. $K = 3$. In this way, $3^{7}=2187$ images were obtained per each scenario, for a total of $6561$ images available for the training of the IFDS. 

The resulting dataset has been used to train an IFDS system based on a pre-trained MobileNetV2 \cite{MobileNetV2} on ImageNet \cite{russakovsky2015imagenet} which is used as a feature extractor. The extracted features are reduced by a Global Average Pooling layer which feeds the classifier composed of one dense layer with one neuron per scenario having softmax activation. The classifier is trained for $20$ epochs using Adam \cite{Adam} as optimizer with a learning rate of $10^{-2}$. Finally, the whole network is fine-tuned for $10$ epochs with a learning rate of $10^{-5}$.

\begin{figure}
\centering
\includegraphics[width=\columnwidth]{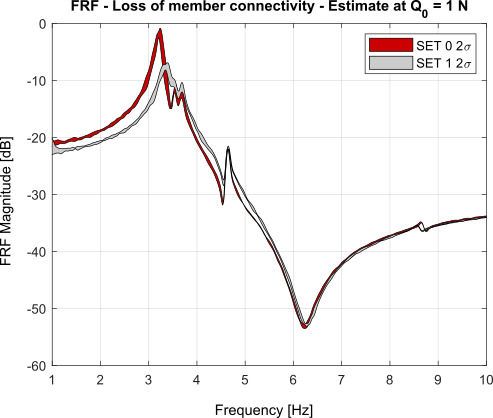}
\caption[caption]{The FRFs of two different acquisition sets of the loss of member connectivity scenario at $Q_{0}$ = 1 N. A clear cluster effect is shown due to changes of frictional conditions and joints play.}
\label{fig:cluster_effect}
\end{figure}

To include experimental data variability, tests have been performed with the same modalities but on different days. Such an additional response variation was crucial to preliminary assess the robustness of the IFDS. Indeed, due to the combined effects of friction and play within the pinned joints, the acquisition sets estimated on different days, after restarting the equipment, brought out a cluster effect. As an example, in Fig. \ref{fig:cluster_effect}, we plot the confidence band ($\pm 2 \sigma$) of the FRF estimated ($Q_{0}$ = 1 N) in the scenario of loss of member connectivity along two sets, acquired on different days. Especially in the low-frequency region, it is clear how the variability induced by resetting the experiment is higher than the variance along consecutive measurements.

\begin{figure}
\centering
\includegraphics[width=0.5\textwidth]{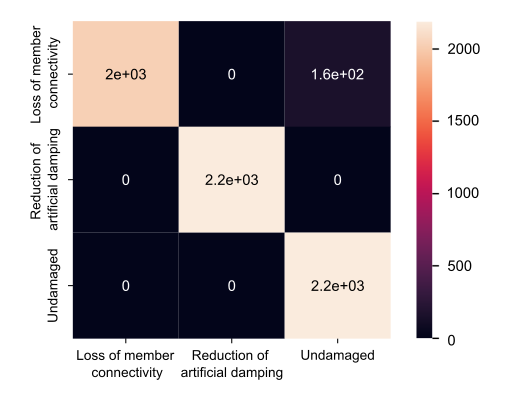}
\caption[caption]{Confusion matrix of the designed IFDS computed on the test set.}
\label{fig:confusion_matrix}
\end{figure}

The images obtained after applying the proposed visualization and augmentation techniques on the new FRFs acquired were fed to the IFDS, obtaining a test accuracy of 97.6\%. Figure \ref{fig:confusion_matrix}, shows the corresponding confusion matrix. As can be seen, the IFDS struggles in correctly classifying some bolt damages which are actually classified as undamaged. Nevertheless, the designed IFDS has been able to overcome the cluster effect shown in Fig. \ref{fig:cluster_effect}, providing empirical support for the real-world application of the proposed method.

\section{Conclusions} \label{sec:conclusions}
This document proposed a novel approach to tackle the problem of data scarcity often met during the design of IFDS. A multi-excitation level procedure leveraging intrinsic non-linearities of real-world systems is used to acquire and augment the available data, producing images that can be conveniently analysed by pre-trained Convolutional Neural Networks (CNNs) to diagnose faults. The proposed approach has been applied to designing an IFDS for the structural health monitoring of a real-scale railway pantograph. Three scenarios were studied: undamage structure, a faulty bolted connection, and a faulty damper.

The designed IFDS was able to correctly classify 97.6\% of the testing conditions, looking at the FRFs obtained at different excitation levels, showing a slight trend in misclassifying bolt damages as the undamaged condition. Nevertheless, the designed IFDS has been able to overcome the cluster effect shown by the pantograph's data acquired at different times, providing empirical support for the real-world application of the proposed method.

\section*{Statements and Declarations} \label{sec:declarations}

\subsection*{Acknowledgments}
The Authors would like to thank Trenitalia SpA for the support given during the developments of the research activities.

\subsection*{Conflict of interest}
The Authors declare that they have no conflict of interest. 

\subsection*{Funding}
The Authors declare that no funds, grants, or other support were received during the preparation of this manuscript.

\subsection*{Authors contribution}
G.S. and A.M.G. equally contributed to the conceptualization of the method, the writing of the manuscript, and the editing of the figures. In particular, G.S. focused on the validation experiments while A.M.G. developed the software routine. M.S. and A.F. supervised the research activity. 

\subsection*{Data availability}
The datasets generated and analysed during the current study are available from the corresponding Author on reasonable request.

\bibliographystyle{plainnat}
\bibliography{bibliography}

\end{document}